\pdfoutput=1
\documentclass[letterpaper]{article} % DO NOT CHANGE THIS
\usepackage{aaai2026}  % DO NOT CHANGE THIS
\usepackage{times}  % DO NOT CHANGE THIS
\usepackage{helvet}  % DO NOT CHANGE THIS
\usepackage{courier}  % DO NOT CHANGE THIS
\usepackage[hyphens]{url}  % DO NOT CHANGE THIS
\usepackage{graphicx} % DO NOT CHANGE THIS
\usepackage{natbib}  % DO NOT CHANGE THIS AND DO NOT ADD ANY OPTIONS TO IT
\usepackage{caption} % DO NOT CHANGE THIS AND DO NOT ADD ANY OPTIONS TO IT
\usepackage{algorithm}
\usepackage{algorithmic}
\usepackage{amsmath}    % For align, gather, etc.
\usepackage{amssymb}    % For \mathbb{R}, etc.
\usepackage{booktabs}
\usepackage{multirow}
\usepackage{newfloat}
\usepackage{listings}
\DeclareCaptionStyle{ruled}{labelfont=normalfont,labelsep=colon,strut=off} % DO NOT CHANGE THIS
\floatstyle{ruled}
\newfloat{listing}{tb}{lst}{}
\floatname{listing}{Listing}
\newcommand{\flux}{{\sf Flux.1-Dev}~}

\title{Taming Identity Consistency and Prompt Diversity in Diffusion Models via Latent Concatenation and Masked Conditional Flow Matching}
\author{
Aditi Singhania\textsuperscript{\rm 1},
Arushi Jain\textsuperscript{\rm 1},
Krutik Malani\textsuperscript{\rm 1},
Riddhi Dhawan\textsuperscript{\rm 1},
Souymodip Chakraborty\textsuperscript{\rm 1},
Vineet Batra\textsuperscript{\rm 1},
Ankit Phogat\textsuperscript{\rm 1}
}
\affiliations{
\textsuperscript{\rm 1} Adobe  \\
}

\nocopyright
\begin{document}

\maketitle

\begin{abstract}
Subject-driven image generation aims to synthesize novel depictions of a specific subject across diverse contexts while preserving its core identity features. Achieving both strong identity consistency and high prompt diversity presents a fundamental trade-off. We propose a LoRA fine-tuned diffusion model employing a \emph{latent concatenation} strategy, which jointly processes reference and target images, combined with a \emph{masked Conditional Flow Matching (CFM)} objective. This approach enables robust identity preservation without architectural modifications. To facilitate large-scale training, we introduce a two-stage \textit{Distilled Data Curation Framework}: the first stage leverages data restoration and VLM-based filtering to create a compact, high-quality seed dataset from diverse sources; the second stage utilizes these curated examples for parameter-efficient fine-tuning, thus scaling the generation capability across various subjects and contexts. Finally, for filtering and quality assessment, we present \textsc{CHARIS}, a fine-grained evaluation framework that performs attribute-level comparisons along five key axes: identity consistency, prompt adherence, region-wise color fidelity, visual quality, and transformation diversity.
\end{abstract}

\section{Introduction}

Text-to-image diffusion models have recently achieved remarkable progress, enabling the synthesis of high-quality and visually diverse images from natural language descriptions~\cite{fluxGithub, esser2024rectifiedflow, chen2024pixartsigma}. As these models evolve, subject-driven customization has emerged as a compelling frontier—enabling the consistent rendering of specific subjects or objects across varied prompts, scenes, and contexts. This capability typically requires conditioning on both textual prompts and a reference image to preserve subject identity. The core challenge in subject-driven generation lies in striking an optimal balance between preserving the subject's identity while enabling rich, prompt-guided visual variations.

Despite advances in general image synthesis, subject-driven generation presents distinct challenges. Models often fail to preserve identity under significant transformations such as changes in pose, viewpoint, or environment, or they adapt inadequately to the prompt. The problem becomes particularly difficult when attempting to maintain fine-grained identity features (e.g., specific facial characteristics, clothing details, or stylistic elements) while allowing for natural variations in expression, posture, or contextual setting.

Existing approaches to subject-driven generation typically fall into three categories. Traditional fine-tuning methods like DreamBooth~\cite{ruiz2022dreambooth, fan2024dreamboothpp} modify model weights for each subject though they need multiple reference images and incur significant computational costs. Encoder-based approaches extract identity features using specialized adapters but often struggle with diverse subjects beyond faces. Recent diffusion transformer (DiT) methods leverage in-context learning capabilities but introduce architectural complexity and lack comprehensive data pipelines necessary for capturing diverse poses, styles, and semantic contexts.

We present a comprehensive framework for identity-preserving image generation. Our approach introduces three key innovations:
\begin{itemize}
\item \textbf{Latent Concatenation Strategy:} Rather than using parallel branches with modified positional embeddings, we propose a simpler but more effective approach that concatenates the noisy target latent with a noise-free reference latent. Combined with a masked Conditional Flow Matching (CFM) objective and LoRA fine-tuning on key transformer layers, this design enables robust identity transfer without requiring architectural modifications to the underlying diffusion model.
\item \textbf{Distilled Data Curation Framework:} We introduce a two-stage pipeline to address data scalability challenges. The first stage creates a small, high-quality seed dataset through data restoration and VLM-based filtering. The second stage uses these curated examples for parameter-efficient fine-tuning to scale generation across diverse subjects and contexts, overcoming the limitations of previous methods that relied on smaller, less diverse datasets.
\item \textbf{CHARIS Evaluation Metric:} We develop a fine-grained evaluation framework that performs attribute-level comparisons across five key axes: identity consistency, prompt adherence, region-wise color fidelity, visual quality, and transformation diversity. Unlike previous metrics that provide high-level assessments, CHARIS offers detailed insights into model performance across different dimensions of identity preservation.
\end{itemize}
Experimental results demonstrate that our approach significantly outperforms existing methods in maintaining subject fidelity under diverse generative conditions while preserving prompt adherence and generation quality.

\section{Related Work}

\subsection{Subject-Driven Generation}
Subject-driven generation aims to synthesize novel depictions of a given subject under varying prompts while preserving core identity features. Multiple paradigms have emerged to address this challenge, each with distinct advantages and limitations.
\paragraph{Per-Subject Fine-Tuning Methods.}
Early approaches to subject-driven generation focused on model fine-tuning. DreamBooth~\cite{ruiz2022dreambooth} pioneered this approach by associating subjects with rare tokens and fine-tuning the diffusion model on specific instances. Textual Inversion~\cite{gal2023textualinversion} optimized text embeddings rather than model weights, while Custom Diffusion~\cite{kumari2022multiconcept} applied more efficient parameter updates. These methods achieved strong identity preservation but suffered from high computational overhead (requiring 15-30 minutes per subject), limited generalization to novel contexts, and the need for multiple high-quality reference images.
\paragraph{Encoder-Based Approaches.}
To avoid costly per-subject finetuning, encoder based approaches have been proposed as a more efficient alternative. 
IP-Adapter~\cite{ye2023ip} projects CLIP visual features into the text-embedding space, enabling plug-and-play personalization. 
BLIP-Diffusion~\cite{li2024blip} and Elite~\cite{wei2023elite} harness pretrained multimodal encoders to capture subject appearance, whereas SSR-Encoder~\cite{zhang2024ssr} improves spatial alignment through token-to-patch matching. Encoders specialized for faces have proved especially effective at preserving identity. 
PhotoMaker~\cite{li2024photomaker}, InstantID~\cite{wang2024instantid}, FaceStudio~\cite{yan2023facestudio}, and PULID~\cite{guo2024pulidpurelightningid} build upon dedicated face recognition backbones, while UniPortrait~\cite{he2024uniportrait} and AnyStory~\cite{he2025anystory} incorporate routing mechanisms for sharper subject localization. 
Nevertheless, because their encoders remain domainspecific, these methods often perform poorly when handling subjects such as animals, inanimate objects, or stylized characters, and they can miss fine grained details that lie outside their training distribution.
\paragraph{DiT-Based Methods.}
Recent DiT-based methods such as OminiControl~\cite{tan2025ominicontrol}, UNO~\cite{wu2025lesstomore}, and DSD~\cite{cai2025dsd} leverage the in-context generation capabilities of diffusion transformers—particularly those based on the Flux.1 architecture—to mitigate data limitations and achieve identity preservation without explicit supervision. These models typically incorporate parallel conditioning branches along with positional biases, spatially shifting condition tokens to avoid overlap with noisy image tokens. While architecturally leaner than UNet-based systems with specialized encoders, they still introduce structural modifications and require careful coordination of token positions.
Most DiT-based methods begin by generating synthetic training pairs through in-context composition. While effective, these strategies introduce architectural complexity and lack scalable data pipelines with sufficient diversity in pose, style, and semantics.
In contrast, our method performs reference latent conditioning directly in the DiT latent grid using a simple concatenation strategy and applies a masked Conditional Flow Matching loss. This enables scalable and disentangled identity control without modifying the underlying architecture or relying on auxiliary modules. 
\subsection{Evaluation Metrics for Identity Consistency}
Traditional metrics like CLIP~\cite{radford2021learning}, DINOv2~\cite{oquab2024dinov2}, and FID~\cite{heusel2017gans} provide global similarity measures but fail to capture fine-grained identity traits essential to subject-driven generation, especially during complex transformations involving pose or viewpoint changes. Recent benchmarks like DreamBench++~\cite{peng2025dreambenchpp} employ VLMs for more human-aligned evaluation, yet still offer primarily high-level assessments lacking granularity over identity-specific attributes.
Moreover, current VLMs exhibit limitations in detecting subtle visual discrepancies—such as minor color shifts or localized design alterations—which are critical for character fidelity. These nuances often go unnoticed in global assessments, leading to overestimated identity scores. \textbf{CHARIS} (Character Holistic Attribute-Ranked Identity Scoring) builds upon these advances with a structured, multi-axis evaluation protocol across five orthogonal dimensions: identity consistency, prompt adherence, region-wise color fidelity, visual quality, and transformation diversity.
CHARIS incorporates attribute-level comparison and region-specific color analysis in multiple color spaces, enabling interpretable and granular identity assessment. Our approach integrates both VLM and non-VLM techniques to leverage the strengths of each method, providing a more comprehensive evaluation framework that better aligns with human perception of identity consistency to improve on the prior methods.

\section{Method}
\subsection{Preliminaries}
\label{sec:preliminaries}
Recent advancements in generative modeling have introduced transformer-based denoising networks, notably the Diffusion Transformer (DiT) architecture. DiT replaces the traditional U-Net backbone with a pure transformer model, enabling efficient and scalable diffusion processes in the latent space \citep{peebles2023dit}, where the latent space defined by the encoding space of VAE (as introduced in~\cite{rombach2022latentdiffusion}). 
This architecture forms the foundation of state-of-the-art latent diffusion models such as FLUX.1 and Stable Diffusion 3 \citep{fluxGithub, esser2024rectifiedflow}.

At each denoising timestep, the model processes two primary inputs:  
(1) \emph{Latent image tokens} \( z \in \mathbb{R}^{H \times W \times d} \), representing the noisy latent representation of the image, where \( H \) and \( W \) denote the height and width of the token grid, respectively, and \( d \) is the embedding dimension;  
(2) \emph{Text condition embeddings} \( c_T \in \mathbb{R}^{M \times d} \), derived from the textual prompt, with \( M \) being the number of text tokens.  

For processing within the transformer, the 2D latent image tokens are flattened into a sequence \( z' \in \mathbb{R}^{H \cdot W \times d} \). These image tokens are then concatenated with the text condition embeddings to form the combined input sequence:  
\[
x = [z'; c_T] \in \mathbb{R}^{(H \cdot W + M) \times d}
\]

To encode spatial information in the flattened image sequence, \emph{Rotary Position Embeddings} (RoPE) 
% ~\citep{su2021roformer} 
are applied. Given query and key projections \( q_{i,j} = W_q x_{i,j} \) and \( k_{i,j} = W_k x_{i,j} \), RoPE transforms these projections based on spatial positions \((i,j)\):
\begin{equation}\label{eq:rope}
q'_{i,j} = \text{RoPE}(q_{i,j}, \omega_{i,j}), \quad k'_{i,j} = \text{RoPE}(k_{i,j}, \omega_{i,j})    
\end{equation}

where \(W_q, W_k \in \mathbb{R}^{d\times d}\) are learnable weight matrices, and \(\omega_{i,j}\) encodes sinusoidal positional frequencies. Text embeddings lack spatial extent and are assigned a default position \((0,0)\).
The combined tokens undergo multi-modal attention (MMA), facilitating cross-modal interactions:
\[
\text{MMA}(x) = \text{softmax}\left(\frac{QK^\top}{\sqrt{d}}\right)V,
\]
where \( Q, K, V \) represent stacked query, key, and value matrices, respectively. MMA ensures semantic coherence between the latent image and textual conditions, enabling controlled image generation.

\begin{figure}[t]
\centering
\includegraphics[width=\linewidth]{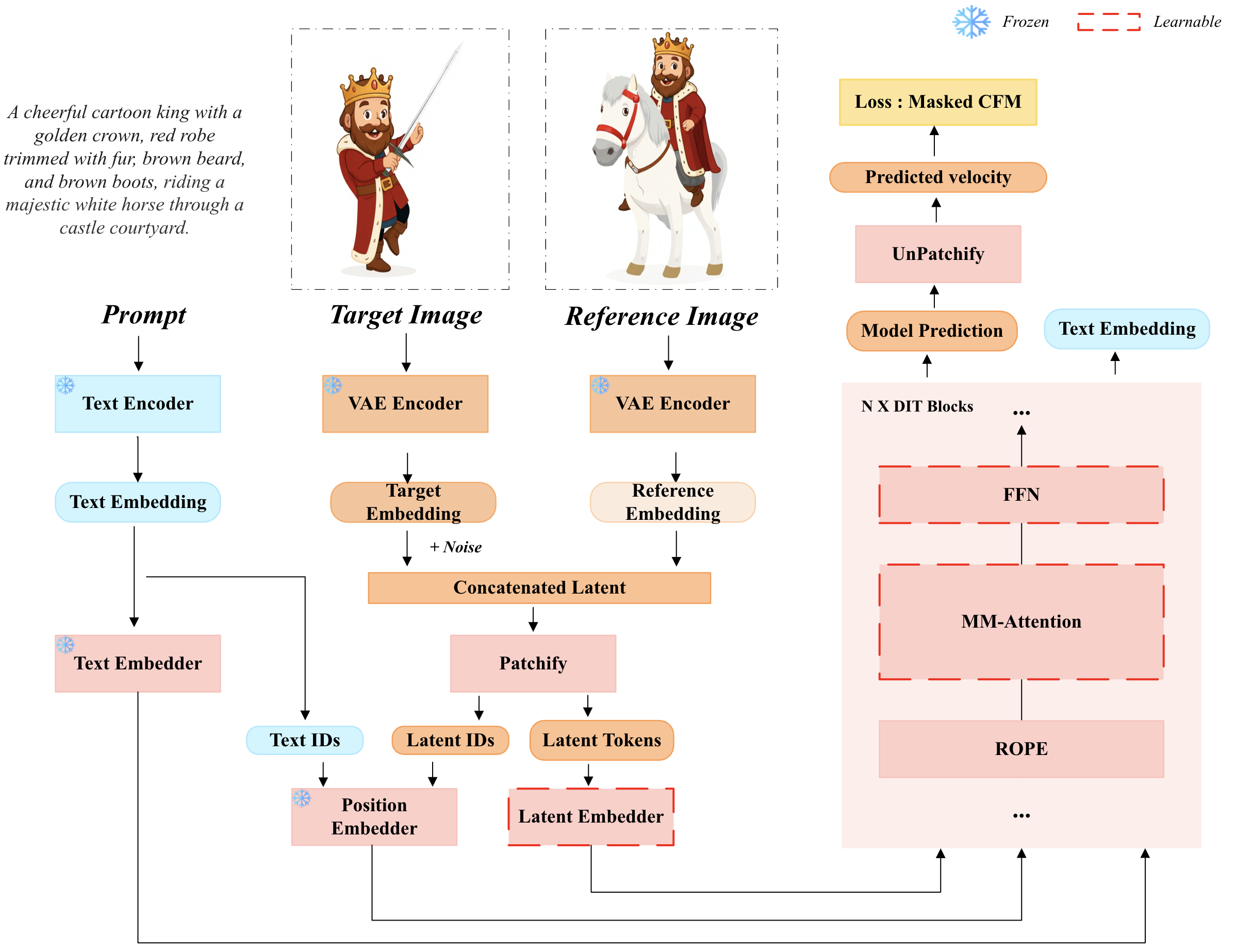}
\caption{Overview of our reference-conditioned diffusion model architecture.}
\label{fig:model_architecture}
\end{figure}

\subsection{Model Architecture: \textsc{Instruct Identity}}
\label{sec:instruct-identity}
We extend the FLUX.1 architecture by introducing an augmented latent input and a masked denoising objective to enable effective identity control without modifying the core transformer structure.

\paragraph{Concatenated Latent Input for Identity Transfer.}
Given a latent reference image \( z_{\text{ref}} \in \mathbb{R}^{H \times W \times d} \), our goal is to generate a latent target image \( z_{\text{tgt}} \in \mathbb{R}^{H \times W \times d} \) for the specific text condition $c_T$. We achieve this by modeling the diffusion process from isotropic noise \( z_1 \in \mathbb{R}^{H\times 2W \times d} \) towards a concatenated latent \( z_0 \in \mathbb{R}^{H\times 2W \times d} \), formed by stacking \( z_{\text{tgt}} \) and \( z_{\text{ref}} \) along the width dimension.

Spatial positions \((i,j)\) within this combined latent are encoded using Rotary Position Embeddings (RoPE), applied directly to the attention queries and keys as described in equation~\eqref{eq:rope}. Consider an illustrative example with latent dimensions \( H = 4 \), \( W = 4 \), and patch size \(2 \times 2\). Concatenating target and reference latents produces \( z \in \mathbb{R}^{4\times 8 \times d} \), where indices \( j=0 \) to \( 3 \) correspond to the target region, and indices \( j=4 \) to \( 7 \) correspond to the reference region. The relative positional encoding provided by RoPE influences attention across target and reference tokens.
% This concatenation strategy maintains spatial separation while enabling joint processing. 
Furthermore, incorporating text embeddings $c_T$ within the same attention sequence provides semantic grounding, ensuring identity-preserving generation guided by textual prompts.

To ensure that updates during training occur exclusively within the target region, we introduce a binary spatial mask \(\mu\) applied to the latent space. The mask selectively activates gradients over the left half (target region), defining the conditional flow matching (CFM)~\citep{lipman2023flowmatching} objective with masked loss:
\begin{equation}
\mathcal{L}_{\text{CFM}}^{\mu} = \mathbb{E}_{t \in \mathcal{U}} \left[ \left\| \mu \cdot \left( z_0 - z_1 - v_\theta(z_t; t; c_T) \right) \right\|^2 \right],
\end{equation}
where \(v_\theta\) denotes the velocity flow field prediction model, and the spatial mask \(\mu\) is defined as:
\begin{equation*}
\mu_{i,j} =
\begin{cases}
1 & \text{if } j < W, \\[3pt]
0 & \text{otherwise.}
\end{cases}
\end{equation*}
This masking strategy confines gradient propagation to parameters directly influencing the transformation of the target region, while the reference region serves as a fixed identity anchor, enhancing identity-preserving capabilities.
\subsection{Distilled Data Curation Framework}
\label{sec:data_curation}
A significant challenge in subject-driven image generation is obtaining diverse, high-quality training data that maintains consistent identity across varied contexts. While large text-to-image models can theoretically supply this data \citep{huang2024incontextlora, cai2025dsd}, their outputs often lack necessary quality at scale due to: (1) poor layout awareness causing improper composition, (2) cross-region inconsistency leading to identity drift, (3) reduced prompt adherence limiting control, (4) incomplete outputs with missing elements, and (5) limited scenario diversity.
\begin{figure}[t]
  \centering
  \includegraphics[width=\columnwidth]{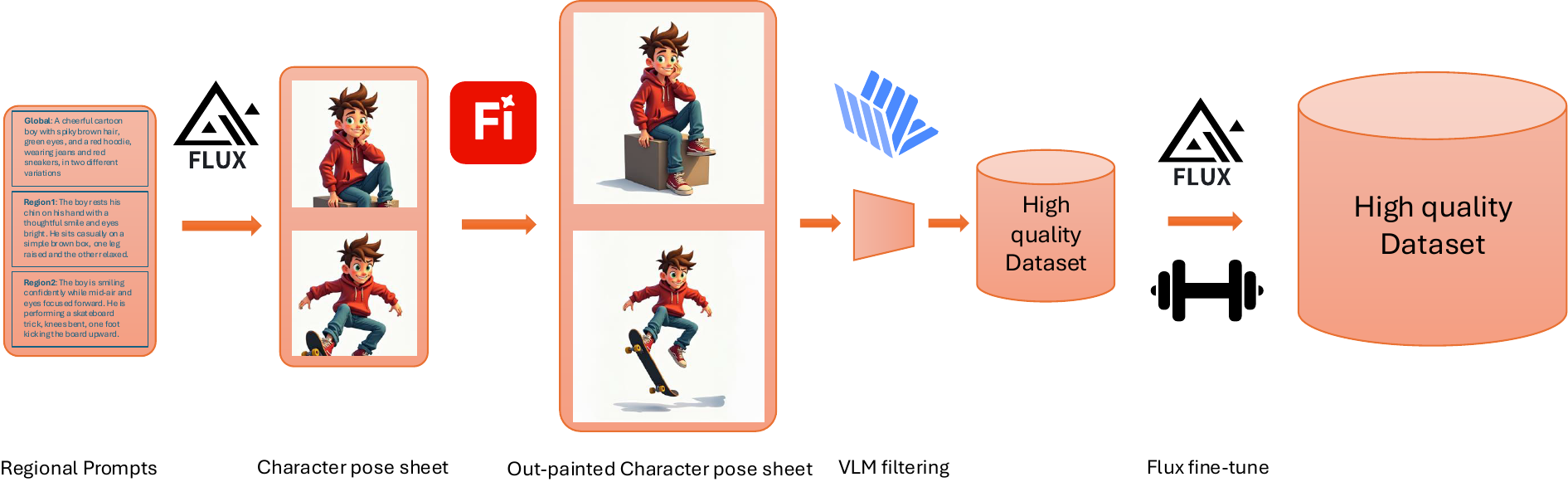}
  \caption{Pipeline for constructing a subject-consistent, high-quality dataset via regional prompting and visual-language model (VLM) filtering. A cartoon boy character is generated with Flux using global and regional prompts, followed by out-painting to extend pose diversity. VLM filtering ensures semantic consistency, and the resulting dataset is used to fine-tune the Flux model for improved subject identity retention across varied contexts.}
  \label{fig:data_curation}
\end{figure}
\subsubsection{Stage I – High-Quality Supervised Dataset Creation}
We address the above limitations of generating high quality data through text-to-image models through three key techniques:
\paragraph{Regional Prompting for Structured Layouts}
To address layout incoherence, cross-region inconsistency, and prompt saturation, we adopt a regional prompting strategy inspired by prior structured generation techniques~\citep{chen2024regionalprompting}. The canvas \( C \) is partitioned into \( R \) disjoint spatial regions \( \{C_1, C_2, \ldots, C_R\} \), each associated with a region-specific prompt \( p_r \) and mask \( M_r \), while a shared identity descriptor \( p_{\text{id}} \) governs visual consistency:
\begin{align*}
    C & = \bigcup_{r=1}^{R} C_r, \quad C_r \cap C_{r'} = \emptyset \ (\forall r \ne r') \\
    \mathcal{P} & = (p_{\text{base}}, \{(p_r, M_r)\}_{r=1}^{R}) 
\end{align*}

This structured conditioning guides the model to generate subject-consistent content within designated regions, forming complete character sheets with controlled variation. However, applying this setup directly to base models introduces significant inference overhead due to sequential region-wise passes and still results in visual failures such as cropped limbs, color inconsistencies, and identity drift. 
\paragraph{Quality Improvement with Generative Out-painting} 
To address incomplete outputs, we apply out-painting to extend backgrounds and recover truncated elements. This process preserves the existing visual content while expanding scene completeness, significantly improving generation quality while maintaining stylistic and identity coherence.
\paragraph{VLM-Based Filtering via CHARIS}
We employ CHARIS (Section~\ref{sec:charis}) to filter candidate pairs using thresholds across multiple quality axes. Only samples passing stringent criteria for identity consistency, prompt adherence, and transformation diversity are retained, ensuring high-quality training data.
\subsubsection{Stage II – Parameter-Efficient Fine-Tuning}
We generate supervised image pairs by fine-tuning a pretrained text-to-image model using parameter-efficient methods. Specifically, we apply LoRA tuning to a minimal subset of attention layers to adapt the model for generating identity-consistent image grids without modifying its core architecture. Compared to base model outputs, the fine-tuned model produces more prompt-aligned generations with reduced identity drift, enabling scalable synthesis of high-quality training data with significantly lower post-processing requirements.

\begin{figure}
    \centering
    \begin{minipage}{0.3\columnwidth}
        \centering
        \includegraphics[width=\textwidth]{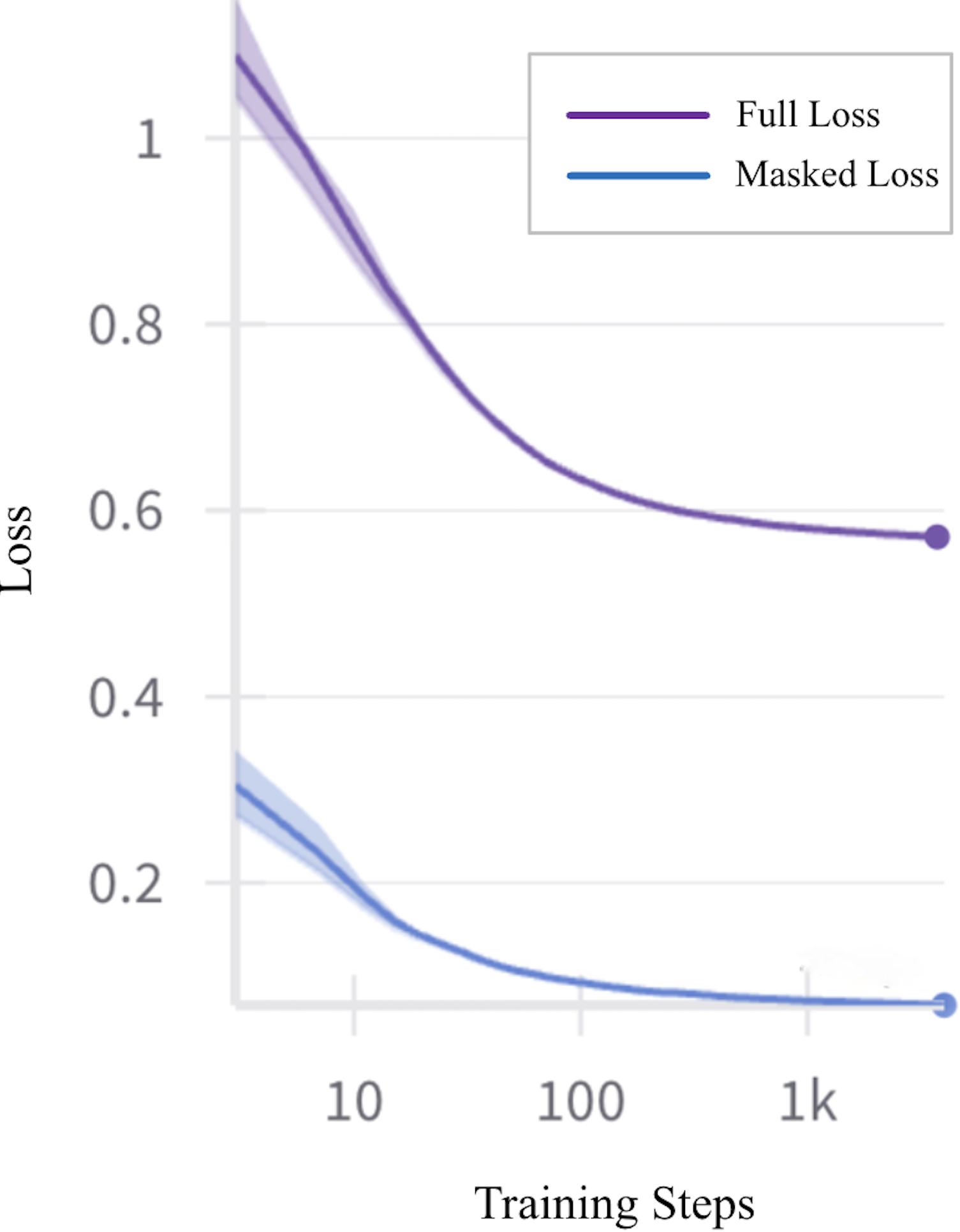}
        \caption{Training masked loss vs full loss.}
        \label{fig:loss_curve}
    \end{minipage}
    \hfill
    \begin{minipage}{0.45\columnwidth}
        \centering
        \includegraphics[width=\textwidth]{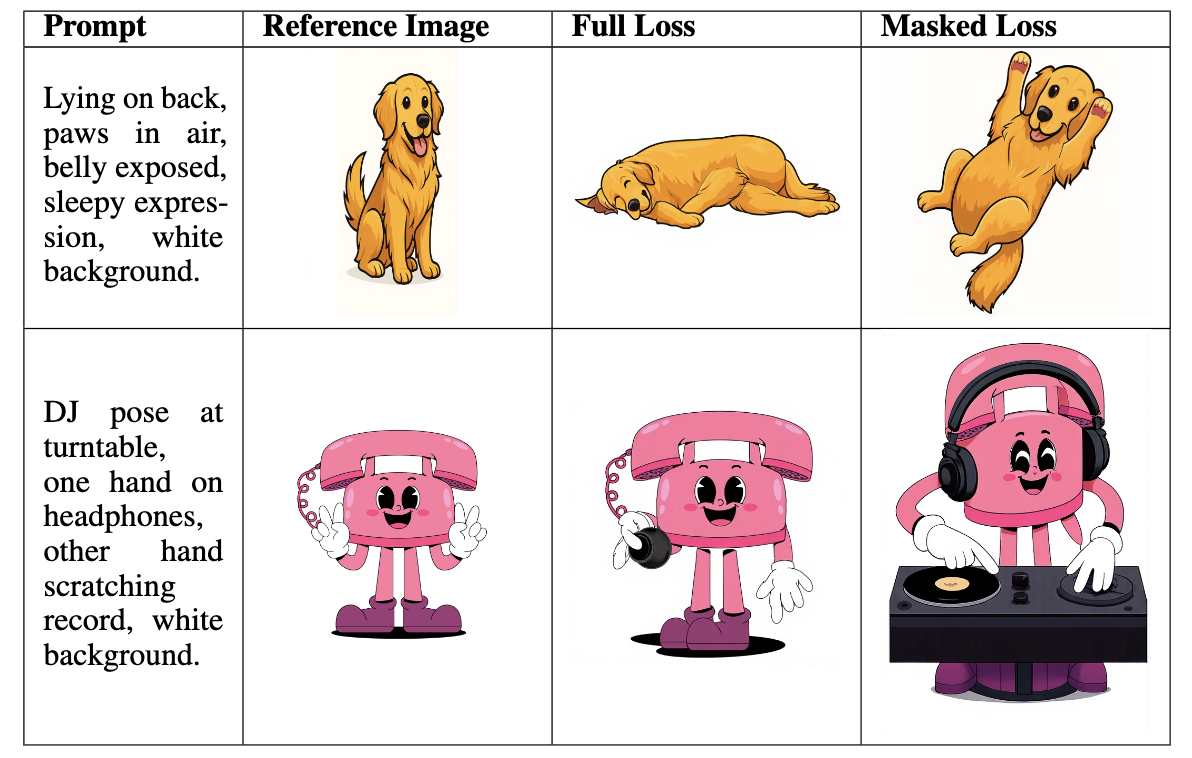}
        \caption{Full loss vs masked loss outputs for given prompts.}
        \label{fig:loss_outputs}
    \end{minipage}
\end{figure}

\subsection{CHARIS : The Evaluation Framework}
\label{sec:charis}
We introduce \textbf{CHARIS (Character Holistic Attribute-Ranked Identity Scoring)}, a training-free and interpretable evaluation metric tailored for subject-driven image generation. CHARIS integrates fine-grained assessments across five key dimensions into a single, human-aligned composite score. Unlike prior metrics~\cite{peng2025dreambenchpp, cai2025dsd} that rely on coarse visual similarity scores, CHARIS provides attribute-level evaluations that more accurately capture identity preservation and align closely with human judgment. 
The framework combines VLM-based attribute matching, prompt understanding, and axis-specific diversity analysis with CLIP-based prompt adherence and perceptual quality. Region-wise color fidelity is computed via segmentation \cite{ren2024groundedsam} and color-space deviation scoring (\textbf{Algorithm~\ref{alg:color}}). The full scoring pipeline is presented in \textbf{Algorithm~\ref{alg:charis}}. CHARIS computes a weighted geometric mean across axes to ensure that poor performance along any single dimension reduces the overall score. We use equal weights ($w_a = 0.25$) for benchmarking and adjust them during data filtering to prioritize fidelity. The implementation details and thresholds are provided in the supplementary material.

\begin{algorithm}[tb]
\small 
\hrule
\caption{ColorScore$(I_{\text{ref}}, I_{\text{gen}})$}
\label{alg:color}
\hrule\medskip
\begin{algorithmic}[1]      % [1] → number every \STATE + control line
  \STATE $R_{\text{ref}} \gets \mathrm{Segment}(I_{\text{ref}})$
  \STATE $R_{\text{gen}} \gets \mathrm{Segment}(I_{\text{gen}})$
  \STATE $M \gets \mathrm{MatchRegions}(R_{\text{ref}}, R_{\text{gen}})$
  \FOR{$(r_1,r_2)\in M$}
    \FOR{$\sigma\in\{\mathrm{RGB},\mathrm{HSV},\mathrm{LAB}\}$}
      \STATE $\delta \gets \lVert\mu(r_1^\sigma)-\mu(r_2^\sigma)\rVert_2$
      \STATE\[
        S^\sigma_{r_1,r_2} =
        \begin{cases}
          0 & \delta \ge t_2^\sigma \quad (\text{inconsistency})\\
          1 & t_1^\sigma \le \delta < t_2^\sigma \quad (\text{deviation})\\
          2 & \delta < t_1^\sigma \quad (\text{consistency})
        \end{cases}
      \]
    \ENDFOR
  \ENDFOR
  \STATE $S_{\text{color}} \gets \dfrac{1}{2|M||\Omega|}\sum S^\sigma_{r_1,r_2}$
  \STATE \textbf{return} $S_{\text{color}}$
\end{algorithmic}
\hrule
\end{algorithm}

%------------------------
% Algorithm 2: CHARIS Evaluation
%------------------------
\begin{algorithm}[tb]         % top or bottom float only
\small 
\hrule
\caption{CHARIS Evaluation Framework}
\label{alg:charis}
\hrule\medskip                % small gap below caption

\begin{algorithmic}[1]        % [1] → number every \STATE + control line
  %-------------------- inputs / outputs --------------------
  \STATE \textbf{Input:} $I_{\text{ref}},\ I_{\text{gen}},\ T,\ \text{mode}$
  \STATE \textbf{Output:} CHARIS score
  %------------------- component scores ---------------------
  \STATE $S_{\text{id}}      \gets \mathrm{VLM\text{-}ID}(I_{\text{ref}}, I_{\text{gen}}, T)/4$
  \STATE $S_{\text{prompt}}  \gets \bigl(\mathrm{VLM\text{-}Prompt}/4 + \mathrm{CLIP}\bigr)/2$
  \STATE $S_{\text{color}}   \gets \mathrm{ColorScore}(I_{\text{ref}}, I_{\text{gen}})$
  \STATE $S_{\text{quality}} \gets \mathrm{CLIP\text{-}IQA}(I_{\text{gen}})$
  \STATE $S_{\text{div}}     \gets \mathrm{VLM\text{-}DiversityScore}(I_{\text{ref}}, I_{\text{gen}}, T)$
  %------------------- weighting scheme ---------------------
  \IF{$\text{mode} = \text{Benchmarking}$}
    \STATE $A \gets \{\text{id},\ \text{prompt},\ \text{color},\ \text{quality}\}$
    \STATE $w \gets w^{\text{bench}}$
  \ELSE
    \STATE $A \gets \{\text{id},\ \text{color},\ \text{quality},\ \text{diversity}\}$
    \STATE $w \gets w^{\text{filter}}$
  \ENDIF
  %------------------- final aggregation --------------------
  \STATE $\text{CHARIS} \gets
         \left( \displaystyle\prod_{a \in A} S_a^{\,w_a} \right)^{\!1 / \sum w_a}$
  \STATE \textbf{return} CHARIS
\end{algorithmic}
\hrule                      % bottom rule
\end{algorithm}

% -----------------------------------  Illustration  -----------------------------------

% -----------------------------------  Main comparison table  --------------------------
\begin{table}[tb]
\small 
  \centering
  \fontsize{9}{10.8}\selectfont                % 9-pt table text (allowed)
  \setlength{\tabcolsep}{2.2pt}
  \renewcommand{\arraystretch}{1.1}
  \begin{tabular}{lccc c}
    \toprule
      \textbf{Metric} & \textbf{UNO} & \textbf{DSD} & \textbf{OmniC} & \textbf{Ours} \\
    \midrule
      $\textbf{CP} \uparrow$                & 0.809 & 0.688 & 0.594 & 0.812 \\
      $\textbf{PF} \uparrow$                & 0.750 & 0.781 & 0.844 & 0.938 \\
      $\textbf{CP}\!\cdot\!\textbf{PF} \uparrow$ & 0.607 & 0.537 & 0.501 & 0.762 \\
    \midrule
      \textbf{Deb. CP} $\uparrow$           & 0.778 & 0.688 & 0.688 & 0.781 \\
      \textbf{Deb. PF} $\uparrow$           & 0.656 & 0.625 & 0.781 & 0.906 \\
      \textbf{Deb. CP$\!\cdot\!$PF} $\uparrow$ & 0.510 & 0.430 & 0.537 & 0.708 \\
    \midrule
      \textbf{CHARIS CP} $\uparrow$         & 0.844 & 0.750 & 0.781 & 0.906 \\
      \textbf{CHARIS PF} $\uparrow$         & 0.617 & 0.610 & 0.617 & 0.662 \\
      \textbf{CHARIS RC} $\uparrow$         & 0.464 & 0.590 & 0.604 & 0.670 \\
      \textbf{CHARIS VQ} $\uparrow$         & 0.693 & 0.747 & 0.747 & 0.788 \\
      $\textbf{CHARIS} \uparrow$            & 0.640 & 0.670 & 0.683 & 0.750 \\
    \bottomrule
  \end{tabular}
  \caption{Quantitative comparison with recent zero-shot baselines. Our approach
           outperforms UNO, DSD and OmniControl on every metric.}
  \label{tab:methods}
\end{table}

% -----------------------------------  Ablation table  ---------------------------------
\begin{table}[tb]
\small 
  \centering
  \fontsize{9}{10.8}\selectfont
  \setlength{\tabcolsep}{2pt}
  \renewcommand{\arraystretch}{1.1}
  \begin{tabular}{lccc cc}
    \toprule
      \multirow{2}{*}{\textbf{Metric}} &
      \multicolumn{3}{c}{\textbf{LoRA Layers}} &
      \multicolumn{2}{c}{\textbf{Loss Type}} \\
      \cmidrule(lr){2-4}\cmidrule(lr){5-6}
      & \textbf{A} & \textbf{B} & \textbf{C} & \textbf{Full} & \textbf{Masked} \\
    \midrule
      $\textbf{CP} \uparrow$                & 0.700 & 0.812 & 0.730 & 0.538 & 0.812 \\
      $\textbf{PF} \uparrow$                & 0.889 & 0.938 & 0.861 & 0.600 & 0.938 \\
      $\textbf{CP}\!\cdot\!\textbf{PF} \uparrow$ & 0.622 & 0.762 & 0.629 & 0.323 & 0.762 \\
    \midrule
      \textbf{Deb. CP} $\uparrow$           & 0.900 & 0.781 & 0.899 & 0.712 & 0.781 \\
      \textbf{Deb. PF} $\uparrow$           & 0.830 & 0.906 & 0.774 & 0.605 & 0.906 \\
      \textbf{Deb. CP$\!\cdot\!$PF} $\uparrow$ & 0.747 & 0.708 & 0.696 & 0.431 & 0.708 \\
    \midrule
      \textbf{CH. CP} $\uparrow$            & 0.903 & 0.906 & 0.901 & 0.712 & 0.906 \\
      \textbf{CH. PF} $\uparrow$            & 0.635 & 0.638 & 0.624 & 0.514 & 0.662 \\
      \textbf{CH. RC} $\uparrow$            & 0.625 & 0.670 & 0.667 & 0.663 & 0.670 \\
      \textbf{CH. VQ} $\uparrow$            & 0.771 & 0.788 & 0.779 & 0.749 & 0.788 \\
      $\textbf{CH.} \uparrow$               & 0.725 & 0.743 & 0.735 & 0.652 & 0.750 \\
    \bottomrule
  \end{tabular}
  \caption{Ablation results. Left: LoRA configurations (A = attention-only,
           B = attention+FFN, C = attention+FFN+modulation). Right:
           comparison of loss functions. Configuration B with masked loss
           performs best.}
  \label{tab:ablation}
\end{table}

\begin{figure*}                            % span two columns, top / bottom only
  \centering
  \includegraphics[width=0.92\linewidth]{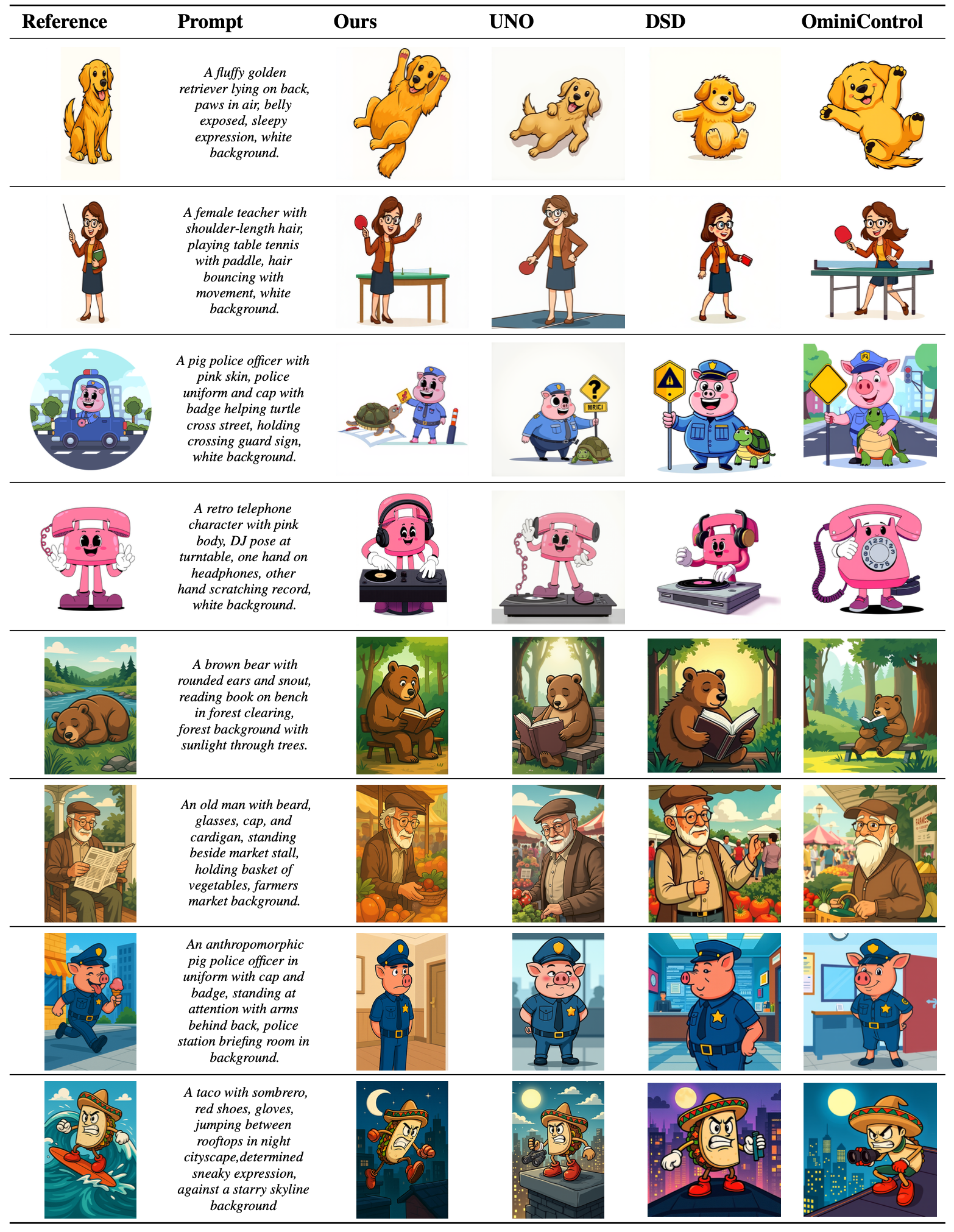}  % one dimension only
  \caption{Visual comparison of identity preservation across diverse subjects.
           Our method maintains character-defining features and colour fidelity
           compared to UNO~\cite{wu2025lesstomore}, DSD~\cite{cai2025dsd},
           and OmniControl~\cite{tan2025ominicontrol}.}
  \label{fig:performance}
\end{figure*}

\section{Experiments}
\paragraph{Implementation Details.} 
We trained our models on $8\times\text{NVIDIA}$ A100 (80GB) GPUs using mixed-precision (FP16). For data curation, we first fine-tuned the base \flux~model using LoRA ($r=16$) for 5,000 steps, creating scalable high-quality datasets. Our \textsc{Instruct Identity} model was trained with LoRA ($r=512$) for 10,000 steps using concatenated latent input and masked Conditional Flow Matching objective. Both phases used AdamW optimizer (lr=1e-4).

\paragraph{Datasets.}

Our automated Data Curation Framework (Section~\ref{sec:data_curation}) generated two datasets: 10,000 consistent image grids for initial refinement and 220,000 consistent identity and diverse scenarios image pairs for \textsc{Instruct Identity} training. The curation pipeline involved InternVL2.5-38-MPO~\cite{chen2024internvl} for high-quality data filtering and captioning, FLUX.1-Fill-dev~\cite{fluxGithub} for generative out-painting and fine-tuning of the base text-to-image model \flux. Image pairs were selected using composite scoring across identity consistency, prompt adherence, transformation diversity, and visual completeness metrics. The final dataset balances humans, animals, anthropomorphic characters, and animated objects.

\paragraph{Evaluation Protocol.}
We benchmark our method against recent zero-shot models, including DSD (\citet{cai2025dsd}), OminiControl~\cite{tan2025ominicontrol}, and UNO~\citep{wu2025lesstomore}. Results are presented in Table~\ref{tab:methods}. While conventional metrics such as CLIP similarity and FID scores are commonly used for image generation tasks, they inadequately capture the nuanced requirements of zero-shot character customization. As noted by \citep{fan2024dreamboothpp}, these metrics tend to be noisy and often favor identity mappings (i.e., minimal transformation), limiting their reliability for our evaluation context. Therefore, we evaluate model performance using CHARIS, which quantifies five dimensions critical to subject-driven generation: identity consistency, prompt adherence, region-level color fidelity, visual quality, and output diversity. To contextualize results, we also compare with evaluation setups from prior works—such as DSD's GPT-4o-based scoring on DreamBench++. In contrast to traditional metrics like CLIP, DINO, or FID, CHARIS provides more detailed, interpretable feedback aligned with human perception. All reported scores are averaged.
\subsection{Qualitative Comparison}
Figure~\ref{fig:performance} presents visual comparisons across diverse scenarios challenging pose complexity, subject categories, and fine-grained visual attributes. Against recent DIT-based methods, \textbf{UNO}~\cite{wu2025lesstomore} shows adequate generation but struggles with color fidelity and facial consistency across viewpoints. \textbf{DSD}~\cite{cai2025dsd} better preserves general identity but exhibits anatomical inaccuracies and reduced background adherence. \textbf{OmniControl}~\cite{fan2024dreamboothpp} demonstrates the most significant identity inconsistencies, modifying key character attributes and distorting features under complex prompts. Our \textsc{Instruct Identity} maintains consistent structural integrity throughout all test cases, preserving color distributions, character-defining features, and anatomical correctness regardless of pose complexity. The results \ref{fig:performance} demonstrate our approach achieves superior identity preservation while maintaining strong prompt adherence across all evaluation scenarios.

\subsection{Quantitative Comparison}
Based on the evaluation metrics described in Section~\ref{sec:charis}, we compare our method against recent zero-shot baselines—UNO, DSD, and OmniControl—using DreamBench++, its debiased extension, and our proposed CHARIS framework. As shown in Table ~\ref{tab:methods}, our model achieves superior performance across all consistency axes. While UNO shows strong raw Concept Preservation (CP), our method retains high CP even under debiased settings, indicating reduced copy-paste effects and improved diversity. At the same time, we achieve strong Prompt Following (PF) and Debiased PF, demonstrating robust prompt adherence. Additionally, our method attains the highest regional color consistency and the best overall \textbf{CHARIS} composite score, highlighting its ability to preserve identity while enabling prompt-driven diversity without relying on duplication.

% \begin{figure}[t]
%     \centering
%     \includegraphics[width=0.3\linewidth]{images/graph.png}
%     \caption{Training loss comparison between masked and full objectives. The masked loss (blue) achieves significantly lower values throughout training compared to the full loss objective (purple), demonstrating the effectiveness of our spatial masking approach. Shaded regions represent 95\% confidence intervals across 5 training runs.}
%     \label{fig:masked_vs_full}
% \end{figure}

\subsection{Ablation Studies}
To systematically analyze the impact of our design choices, we conduct detailed ablation experiments on three critical components:
% \paragraph{Effectiveness of Latent Conditioning}
\paragraph{Importance of Spatial Loss Masking}
Since our objective is to transfer identity rather than reconstruct the reference image, we apply a spatial loss mask that excludes the reference region from the denoising objective. This encourages the model to treat the reference image purely as a conditioning signal, aligning the training loss with the desired generation behavior. We compare this against a baseline where the loss is computed over both reference and target regions. As shown in Table~\ref{tab:ablation}, masking significantly improves identity preservation, prompt adherence, and generation diversity. In contrast, full loss introduces identity leakage, reduces diversity, and often results in a copy-paste effect with visual artifacts (see figure~\ref{fig:loss_outputs}). While the full-loss baseline converges faster (see Figure~\ref{fig:loss_curve}), it overfits to the reference, leading to poor generalization. 
These results highlight the importance of aligning the training loss with the identity transfer objective through spatial masking.

\paragraph{LoRA Configuration Study}
Given our latent concatenation strategy (noisy + noise-free image tokens), we consistently apply LoRA to the image embedder across all runs. We ablate three transformer-layer configurations: (i) adapting only attention layers (Config A), (ii) adapting attention and feedforward networks (FFNs) (Config B), and (iii) adapting attention, FFNs, and modulation layers (Config C). As shown in Table ~\ref{tab:ablation}, the attention+FFN configuration yields the best overall performance across subject categories. Adding FFNs improves compositional reasoning and attribute control over attention-only tuning, while further including modulation layers leads to overfitting at longer training steps. These findings highlight that FFN paths are crucial for compositional reasoning, and that jointly adapting attention and FFNs provides the best balance between expressivity and stability.
\section{Conclusion}
We present \textbf{Instruct Identity}, a parameter-efficient reference-conditioned diffusion framework that enables identity-preserving generation across diverse prompts without architectural modifications. Our approach integrates latent concatenation with a spatially masked Conditional Flow Matching objective, ensuring disentangled identity transfer while requiring minimal additional parameters. The introduced two-stage data curation pipeline—establishing a high-fidelity seed dataset followed by lightweight LoRA adaptation—facilitates scalable training for identity-preserving generation. Furthermore, we propose \textbf{CHARIS}, a comprehensive evaluation metric that quantifies identity fidelity, prompt adherence, visual quality, and generative diversity. Empirical evaluations demonstrate that our method achieves superior performance across challenging identity preservation and variation scenarios.
\subsection{Future Improvements}
Our current implementation focuses primarily on single-subject identity preservation. Addressing multi-subject conditioning, especially when characters belong to the same semantic category, and improving robustness to partial or occluded references represents an important direction for future research. Additionally, extending our framework to support alternative control modalities such as sketches, edge maps, or pose keypoints would significantly expand its applicability across diverse computer vision tasks.

\bibliography{aaai2026}


\end{document}